\documentclass[10pt,twocolumn,letterpaper]{article}

\usepackage{graphicx}
\usepackage{times}
\usepackage{epsfig}
\usepackage{graphicx}
\usepackage{amsmath}
\usepackage{amssymb}
\usepackage [english]{babel}
\usepackage [autostyle, english = american]{csquotes}
\MakeOuterQuote{"}
\usepackage{hyperref}

\newcommand{\etal}{et al. }

\usepackage{mwe}
\usepackage[markcase=noupper
]{scrlayer-scrpage}
\ohead{}
\begin{document}

\title{A Useful Taxonomy for Adversarial Robustness of Neural Networks}

\author{Leslie N. Smith \\
	Naval Center for Applied Research in Artificial Intelligence\\
	U.S. Naval Research Laboratory \\
	Washington, D.C.  20375 \\
}

\maketitle


\begin{abstract}
Adversarial attacks and defenses are currently active areas of research for the deep learning community.
A recent review paper divided the defense approaches into three categories; gradient masking, robust optimization, and adversarial example detection.
We divide gradient masking and robust optimization differently: (1)  increasing intra-class compactness and inter-class separation of the feature vectors improves adversarial robustness, and (2) marginalization or removal of non-robust image features also improves adversarial robustness. 
By reframing these topics differently, we provide a fresh perspective that  provides insight into the underlying factors that enable training more robust networks and can help inspire novel solutions.
In addition, there are several papers in the literature of adversarial defenses that claim there is a cost for adversarial robustness, or a trade-off between robustness and accuracy but, under this proposed taxonomy, we hypothesis that this is not universal.
We follow this up with several challenges to the deep learning research community that builds on the connections and insights in this paper.
\end{abstract}

\section{Introduction}

With advances in machine learning technology over the past decade, the use of deep neural networks has had great success in computer vision, speech recognition, robotics, and other applications.
Along with these remarkable improvements in performance, the recognition of vulnerabilities has also increased.
As applications of deep neural networks are increasingly being deployed, the security needs of these applications have come to the foreground, especially for safety-required applications (i.e., self-driving vehicles) and adversarial domains where attacks must be anticipated, such as defense applications.

A recent paper provides a comprehensive review of adversarial attacks and defenses \cite{xu2019adversarial} and provides a taxonomy for both the adversarial attacks and defenses.
Pulling on the past literature, this review paper defines adversarial examples as ``inputs to machine learning models that an attacker intentionally designed to cause the model to make mistakes''.
Here, we present a new perspective on adversarial defenses that we believe can provide clarity and inspire novel defenses to adversarial attacks.

The taxonomy of adversarial defense in Xu, \etal \cite{xu2019adversarial} consists of three categories: gradient masking, robust optimization, and adversarial detection.
Gradient masking includes input data preprocessing (i.e., jpeg compression \cite{guo2017countering},  thermometer encoding \cite{buckman2018thermometer}, adversarial logit pairing \cite{engstrom2018evaluating}), defensive distillation \cite{papernot2016distillation}, randomization of the deep neural network models (i.e., randomly choosing a model from a set of models \cite{tramer2017ensemble} or using dropout \cite{dhillon2018stochastic,feinman2017detecting}), and the use of generative models (i.e., PixelDefend \cite{song2017pixeldefend} and Defense-GAN \cite{samangouei2018defense}).
The theme of this diverse set of defenses is to make it more difficult to create adversarial examples and attacks but Athalye, \etal \cite{athalye2018obfuscated} demonstrate that gradient masking techniques are ineffective\footnote{Specifically, they test the defenses at the ICLR 2018 challenge and found that  7 of 9 defenses relied on obfuscated gradients and their attacks successfully circumvent 6 completely, and 1 partially.}.

The second category in this taxonomy is called robust optimization, and it includes the popular defense method of adversarial training \cite{goodfellow6572explaining}, regularization methods that minimize the effects of small perturbations of the input (i.e.,  Jacobian regularization \cite{jakubovitz2018improving}), and provable defenses (i.e., Reluplex algorithm \cite{carlini2017provably}).
Adversarial training is a form of data augmentation where adversarial examples are added to or replace the benign training data.  
Adversarial training is an important defense discussed in the literature, and variations have been proposed, such as ensemble adversarial training where the adversarial examples are computed from a set of pretrained classifiers \cite{tramer2017ensemble}.
Robust optimization includes methods for making deep neural networks behave more robustly to the presence of adversarial perturbations in the input, which is the primary focus of our taxonomy in Section \ref{sec:tax}.

The third category in this review paper is to detect the presence of adversarial examples in the input in order to protect trained classifiers.
That is, one can design a separate model to classify if a sample is benigned or adversarial.
Carlini and Wagner \cite{carlini2017adversarial} rigorously demonstrate that the properties of adversarial examples are not easy to detect.

For our purposes, we consider adversarial robustness to include all approaches for training networks to improve that network's performance on adversarial examples.
We focus primarily on category 2 of the above taxonomy but we also include many of the methods in their category 1.
We propose this new taxonomy on adversarial robustness to  provide insight to the underlying factors that enable training more robust networks.

In addition, there are several papers in the literature of adversarial attacks and defenses that claim there is a cost for robustness, such that greater robustness requires more data \cite{schmidt2018adversarially}, larger model complexity \cite{madry2017towards}, and longer training times.  Furthermore, there are claims of trade-offs between robustness and accuracy \cite{tsipras2018robustness,kurakin2016adversarial}, and even robustness and simplicity \cite{nakkiran2019adversarial}.
There appears to be widespread acceptance of these claims as universal.
Another motivation of our work is to demonstrate that these claims are appropriate only for a subset of existing methods for training in adversarial robustness.

While there are other taxonomies mentioned in other papers, they offer only well-known factors for dividing approaches.  Guo, \etal \cite{guo2017countering} divide the work in adversarial robustness into model-specific strategies (i.e., adversarial training \cite{goodfellow6572explaining}, regularization methods \cite{jakubovitz2018improving}) and model-agnostic methods (i.e., input preprocessing \cite{zhang2019adversarial}).  Zhang, \etal divide adversarial defense into three categories of data preprocessing \cite{guo2017countering}, gradient masking \cite{athalye2018obfuscated}, and adversarial training \cite{goodfellow6572explaining}.  Here we reframe the category of making networks adversarially robust in order to provide a fresh perspective and inspire novel solutions in a way these other taxonomies do not.


\section{Our taxonomy} 
\label{sec:tax}

There have been several recent papers showing that using metric learning loss functions during training helps in making neural networks more robust to adversarial examples \cite{pang2019rethinking,mustafa2019adversarial,mao2019metric}.
Mustafa, \etal \cite{mustafa2019adversarial} used their own variation of the contrastive center-loss \cite{qi2017contrastive} that  encourages both intra-class compactness and inter-class separation of the feature vectors or logits, which are the activations from the last hidden layer.
The center loss \cite{wen2016discriminative} is a loss function that encourages the feature vectors for each class to lie close to each other (i.e., it encourages intra-class compactness) and the contrastive center-loss function is a generalization of it that also encourages inter-class separation.
We claim that these works imply a  general factor for adversarial robustness, which can be stated as:

\emph{Category 1: Increasing intra-class compactness and inter-class separation of the feature vectors improves adversarial robustness.}

There are several other papers that can be categorized under Category 1.  
Wu and Yu \cite{wu2019understanding} postulate that the training of deep models decreases the average margin while increasing the minimum margin, and recommend increasing the average margin (i.e., the inter-class separation).
Galloway, \etal \cite{galloway2019batch} suggest that batch normalization is a cause of adversarial vulnerability.
This aligns with Category 1 because batch normalization constrains the magnitude of the feature vectors (i.e., the activations in the next to the last layer, which is input to the fully connected and softmax layers).
Hence, batch normalization limits inter-class separation and therefore it can increase adversarial vulnerability.

It is particularly interesting to note that the  defensive distillation approach \cite{papernot2016distillation} utilizes Category 1.
Defensive distillation uses two networks and modifies softmax by dividing by a temperature $T$, such that $ softmax( Z(\theta, x) /T)$, where $ Z(\theta, x)$ is the feature vector, $x$ is the input sample, and $\theta$ are the network's weights.
In a rigorous paper by Carlini and Wagner \cite{carlini2016defensive}, they describe the mechanism behind defensive distillation, they state ``When we train a distilled network at temperature T and then test it at temperature 1, we effectively cause the inputs to the softmax to become larger by a factor of T.''
Since the architecture used in defensive distillation does not contain batch normalization\footnote{See the code for \cite{carlini2017towards}, which is available at \url{https://github.com/carlini/nn_robust_attacks}}, the average magnitude for the feature vectors increases by T, thereby increasing the inter-class separation.
Based on their analysis, we hypothesis that the teacher network (even without distillation) will also show signs of robustness and that adding batch normalization to the architecture or using a feature based attack \cite{sabour2015adversarial} will break the effectiveness of defensive distillation.
We leave testing these speculations as future work.

Additionally, there are a number of papers in the literature focused on improving generalization (but not robustness) by increasing intra-class compactness and inter-class separation of the feature vectors, such as center-loss \cite{wen2016discriminative}, contrastive center-loss \cite{qi2017contrastive}, and lifted structures \cite{oh2016deep}, as well as papers that have appeared recently, such as G-Softmax \cite{luo2019g} and Softmax dissection \cite{he2019softmax}.
Category 1 implies that these methods will improve both generalization and robustness, and we leave testing this as future work.

However, improving both generalization and robustness appears to contradict the conjecture of papers in the literature that suggest there is a trade-off between test accuracy and adversarial robustness \cite{tsipras2018robustness,kurakin2016adversarial}.
This implies the existence of at least one other Category of adversarial robustness where this might be true.
One possible set of defenses include image preprocessing \cite{guo2017countering,zhang2019adversarial} and gradient masking methods (see \cite{athalye2018obfuscated}).  Image preprocessing approaches are based on reducing or eliminating ``non-robust'' adversarial perturbations in the training images.

Adversarial perturbations were described as ``non-robust features'' by Ilyas, \etal \cite{ilyas2019adversarial}.
Ilyas, \etal postulate that machines use all the image features that are discriminatory between classes (assuming the task is classification), even those features that are invisible to humans.
Adversarial training \cite{szegedy2013intriguing} specifically includes training images with non-robust features (i.e., adversarial examples) in order for the network to learn to classify examples with non-robust features properly.

We too believe as described in  Ilyas, \etal \cite{ilyas2019adversarial} that humans and machines perform tasks differently.
For example, humans are limited in the number of image features they use in making a decision while machines are much less limited.
Adversarial examples exist where we expect human performance from a machine.
To attain human performance from a machine, we can manually eliminate non-robust features from the training images via preprocessing or make all non-robust image features non-discriminatory with approaches such as adversarial training.

If we consider the network's training, we realize that as it learns, it averages away the non-discriminatory image features as ``nuisance variables''.
This is analogous to computing the marginal probability by summing or integrating the nuisance variables \cite{goodfellow2016deep}.
Hence, using a bit of inductive reasoning, we hypothesis a second Category for adversarial robustness:

\emph{Category 2: Marginalization or removal of non-robust image features improves adversarial robustness.}

Many of the papers on adversarial robustness seem to lie within this Category 2, including adversarial training \cite{szegedy2013intriguing} and methods of gradients masking \cite{athalye2018obfuscated}.
In addition, we show below with a toy example that there is a trade-off between accuracy and adversarial robustness \cite{tsipras2018robustness} for methods that fall under Category 2 (Note: while many of the papers on the trade-off between accuracy and robustness use the adversarial training defense, a similar argument holds for it).

The most obvious way to train a network at human performance levels is to modify the training data to only contain the robust information we want it to use in classification.
One extreme way to eliminate non-robust image features is to preprocess the training and test images with an edge detection algorithm to produce binary edge images.
These edge images commonly display shape information that humans are able to use to recognize objects.
Training a network on edge images results in a highly robust network because all non-visible perturbations have been removed.
However, the performance on benign images is reduced due to a decrease in discriminatory information between classes in the edge detection images relative to the original imagery.
This example demonstrates the trade-off between accuracy and adversarial robustness.
Of course, edge imagery leaves minimal discriminatory information and there is a range of preprocessing that falls on the spectrum between human and machine image features, such as low pass filtering (i.e., DFT \cite{zhang2019adversarial}), denoising, sparse coding, synthetic imagery, and jpeg compression \cite{guo2017countering}.
Note that it is possible to create examples that can fool even a network trained on edge examples by making large visible changes to the input, but the current definitions of adversarial examples include making small imperceptible changes.

To the best of our knowledge, most of the methods in the literature for attaining adversarial robustness fall under Category 2.
The goal of these methods is to marginalize the non-robust features.
This explains why training on more data improves the adversarial robustness of deep networks (i.e., increases the likelihood of non-robust features appearing in different classes to be marginalized away as nuisance variables) \cite{sun2019towards}.
This also explains the added adversarial robustness from Jacobian regularization \cite{jakubovitz2018improving}, where the loss function trains the network to be invariant to small, non-robust features.
It also suggests new methods to obtain adversarial robustness, such as a variant of adversarial training where one adds the same perturbation to images of different classes to make that perturbation non-discriminatory.

An important avenue for future work is to verify our hypothesis of these two Categories by testing some of the new methods for adversarial robustness implied by them.

\section{Discussion}

While we believe that we have presented a few novel connections and insights that we have not seen in the literature, we must still ask if this taxonomy is useful and if so, how.

First, this taxonomy suggests that both robustness and generalization can be improved simultaneously.  
It clarifies that papers declaring there is a trade-off between robustness and accuracy are misleading because the trade-off is not universal.  
We suggest the deep learning community take up the challenge to discover ways to improve both robustness and generalization rather than pursue the current focus of improving robustness at the expense of accuracy.  
Techniques based on metric learning appear to offer performance improvements in both, and other methods may also exist.
Of course, the other side of this challenge is to create new attacks that defeat any new defenses that improve both generalization and robustness.

Second, our paper proposes eliminating non-robust features from the training data so that trained networks learn to only rely on robust image features.
But we don't delineate an optimal way to process images to contain only robust features.
Obviously binary edge detection images are too extreme as they also eliminate many robust image features.
On the other hand, after low pass filtering (i.e., image blurring) non-robust features still remain.
The challenge still remains to discover an ideal preprocessing method or combination of methods.  

Third, new network training methods can be inspired by our analogy of training to marginalization.  
For example, data preprocessing and augmentation can insure that non-robust image features are explicitly present in multiple or all classes to insure that the network treats them as non-discriminatory.
Similarly, marginalization implies that the training methodology in few-shot meta-learning of changing the tasks every iteration creates more universal features that will be beneficial in transfer learning and perhaps in other scenarios.
In addition, the community can investigate better training data combinations that optimally marginalize non-robust features.
There is much additional work to be done in this direction to better understand the theoretical and practical aspects of marginalization.

Fourth, the separation of methods for making networks more robust into two Categories implies that methods from each Category can be productively combined.  The combination of methods from each Category should provide different strengths to a network or an ensemble of networks.
Combine these with the best methods for each of the categories in Xu, \etal \cite{xu2019adversarial}, and one has an ensemble with the potential to make a solid defense.
Unfortunately, the paper with a title ``Ensembles of weak defenses are not strong'' \cite{he2017adversarial} is misleading because in that paper the authors only tested ensembles of defenses that all fall into a single category, such as detectors or our Category 2 above.
He, \etal \cite{he2017adversarial} do mention that their ``adaptive adversarial examples transfer across several defenses'' which might ``explain why ensembling is not an effective approach''.
It is obvious that each defense in an ensemble must provide some strengths that are orthogonal to all the other defenses and an ensemble of many near identical defenses is not useful.

For example, a potential ensemble might include the best adversarial example detector (e.g., Carlini and Wagner \cite{carlini2017adversarial} found the Bayesian uncertainty estimate of Feinman, \etal \cite{feinman2017detecting} to be the strongest of those they tested), as well as a network trained by ensemble adversarial training \cite{tramer2017ensemble}, plus a dropout network that hides the gradient (i.e., Athalye, \etal compare several methods for hiding gradients and found that randomization \cite{dhillon2018stochastic} to be most effective), and networks each from the two Categories in our taxonomy (i.e., one trained with metric learning and another trained on edge detection images, which will force image perturbations to be visible or else they will be eliminated during preprocessing).

We conjecture that a diverse ensemble, with each member offering orthogonal strengths, will be a strictly more powerful defense than any one defense.  
Ablation studies of an ensemble's members can determine if each member adds to the security of the system.
A rigorous analysis of an ensemble's strengths will also identify its remaining weaknesses and further defense efforts can focus on eliminating these weaknesses.

In addition, we hypothesis in this paper that several of the new methods based on metric learning for improving generalization in the literature \cite{luo2019g,he2019softmax} will also improve robustness.  If this is confirmed, there will be numerous other methods in the literature {i.e., \cite{wen2016discriminative,qi2017contrastive}} that will improve robustness but have not been demonstrated yet. 
At this time we admit that there are many items in this paper where we state that these are left for future work and even these items are more than we have time to pursue on our own.  
Hence, we call out to the research community to collaborate with us to investigate some of these items.

\section{Conclusions}

In this paper we expand the area of adversarial robustness into a taxonomy with two categories; Category 1: increasing intra-class compactness and inter-class separation of the feature vectors improves adversarial robustness, and Category 2: marginalization or removal of non-robust image features also improves adversarial robustness.  
This taxonomy permits an understanding of the underlying factors that drive the adversarial robustness of the known methods, and this understanding allows exploring new methods with the same underlying factors.

In addition, we attempt to dispel several potential misunderstandings and set forth several challenges to the deep learning community, such as the discovery of new methods that improve both robustness and generalization.
There are also a number of research items left as future work, such as optimal ways to eliminate non-robust features from the training data via preprocessing or to optimally marginalize non-robust features via training.

We also propose that a diverse ensemble of defenses, with each member offering orthogonal strengths, will be a strictly more powerful approach than any one defense.
An ensemble of defenses should include all the strongest defenses and should be tested against all of the strongest attacks, in order to find the remaining weaknesses.
Then further research on robustness can concentrate on only the remaining holes in the defenses.
 
We also call on researchers to go further with adversarial defense than is typically done today in the literature.
In addition to the challenge of improving both robustness and generalization, researchers can attempt to simultaneously solve multiple other limitations of deep learning, such as reducing the amount of labeled training data needed and creating adaptable networks that learn continuously.

Futhermore, adversarial defenses must go further than working on small imagery such as MNIST and Cifar, which are the most common benchmarks in the adversarial examples literature.
The community seems ready to venture into higher resolution imagery of ImageNet and real world imagery, such as satellite imagery.

Eventually, the research and engineering communities will need to investigate adversarial attacks and defenses in the context of
safety-required applications (i.e., self-driving vehicles) and adversarial domains where attacks must be anticipated, such as defense applications.
It is only in the context of these applications where complete and secure solutions can be discovered.

{\small
	\bibliographystyle{ieee}
	\bibliography{Taxonomy.bib}
}

\end{document}